\documentclass[preprint,5p,times,twocolumn]{elsarticle}

\pdfoutput=1
\usepackage[utf8x]{inputenc}

\usepackage{amsmath,amssymb}
\usepackage{url}
\usepackage{color}
\usepackage{booktabs, multicol, multirow}
\usepackage{subfig}
\usepackage{tikz}
\usetikzlibrary{positioning}
\usepackage[ruled,vlined,linesnumbered]{algorithm2e}

\definecolor{darkgreen}{RGB}{0,120,0}

\journal{}

\begin{document}

\begin{abstract}	
Given the great interest in creating keyframe summaries from video, it is surprising how little has been done to formalise their evaluation and comparison. User studies are often carried out to demonstrate that a proposed method generates a more appealing summary than one or two rival methods. But larger comparison studies cannot feasibly use such user surveys.  Here we propose a discrimination capacity measure as a formal way to quantify the improvement over the uniform baseline, assuming that one or more ground truth summaries are available. Using the VSUMM video collection, we examine 10 video feature types, including CNN and SURF, and 6 methods for matching frames from two summaries. Our results indicate that a simple frame representation through hue histograms suffices for the purposes of comparing keyframe summaries. We subsequently propose a formal protocol for comparing summaries when ground truth is available.
\end{abstract}


\begin{frontmatter}
\title{On the Evaluation of Video Keyframe Summaries using User Ground Truth}
\author[bangor]{Ludmila I Kuncheva\corref{cor}} \ead{l.i.kuncheva@bangor.ac.uk}
\author[bangor]{Iain A D Gunn\fnref{present}} \ead{i.gunn@mdx.ac.uk} \fntext[present]{Present address: Department of Computer Science, Middlesex University, The Burroughs, London NW4 4BT}
\author[bangor]{Paria Yousefi} \ead{paria.yousefi@bangor.ac.uk}

\cortext[cor]{Corresponding Author}
\address[bangor]{School of Computer Science, Bangor University, Dean Street, Bangor, Gwynedd, Wales LL57 1UT, UK}


\end{frontmatter}

\section{Introduction}

Keyframe selection is aimed at summarising video data~\cite{Truong2007}. The summary should be compact, but also diverse and informative for the viewer.
  
While the literature abounds with methods for keyframe selection, surprisingly little has been done towards developing a formal evaluation protocol. The need for such a protocol is widely acknowledged~\cite{Truong2007,Ejaz2013,Furini2010,Khosla2013,LakshmiPriya2014,Lidon2015,Liu2009,Money2008,delMolino2017}. However, at present authors often develop a bespoke experimental set-up in which their proposed method for keyframe selection compares favourably to just one or two alternative methods. (This is beginning to change, as recently, keyframe summaries obtained through different methods have been collated in a publicly available benchmark repository~\cite{DeAvila2011}\footnote{\url{https://sites.google.com/site/vsummsite/results}}). The measures of quality of the keyframe summaries are typically not commensurable across different studies. A particular problem with current evaluation techniques is the lack of comparison to baseline methods. User studies usually demonstrate some percentage improvement achieved by the proposed method against another method, with respect to a chosen criterion such as informativeness or enjoyability. The percentage improvement varies considerably from one study to the next, and it is difficult to assign meaning, let alone statistical significance, to these percentages if different quality measures are used each time. This raises the question of whether the degree of improvement in the summary justifies the effort involved in the design of the new summarisation method. 

The uncertainty is amplified by the lack of large-scale comparisons between keyframe selection methods over large video repositories. The major obstacle in such an endeavour has been the fact that the evaluation of a keyframe summary requires human input at some level, and user studies are expensive. This difficulty can be addressed by obtaining human-made summaries, which we shall call {\em ground truth} summaries. That is, we ask humans to perform the task which the summary algorithms aim to automate, as opposed to asking them to evaluate the automatic summaries directly. Future automatic keyframe summaries can then be evaluated by matching against the existing ground truth collection, rather than requiring a fresh user survey. 

Suppose that $\gamma(A,B|\Theta)$ is a measure of how close keyframe summaries $A$ and $B$ are, and $\Theta$ is a set of parameters of $\gamma$. High values of $\gamma$ are desirable if one of the summaries is seeking to approximate the other. Let $K$ be the evaluated summary, $GT$ be a ground truth summary, and $U$ be a uniform summary (i.e., a set of frames selected from the video at a constant interval) of the video of interest. Our idea relies on the premise that summaries obtained from purposely designed methods are closer to the user preferences (larger $\gamma(K,GT|\Theta)$) than the (context-blind) uniform keyframe selection is. 

We are interested in proposing a good $\gamma$. Alongside proposing a form for $\gamma$, we will seek a parameter set $\Theta$ which tends to maximise $\left(\gamma(K,GT|\Theta)-\gamma(U,GT|\Theta)\right)$ over a suitable selection of ground truths $GT$ and automatic summaries $K$.

In this paper we propose a generic protocol for comparing keyframe summaries with a set of ground truth summaries. The function $\gamma$ we seek will compare the number of successful pairings between elements of the keyframe set under evaluation, and elements of the ground truth. There are four key questions to be answered about this function.  The first three questions will determine the elements of the parameter set $\Theta$:
\begin{enumerate}
	\item {\em Features.} What features should be used to describe the keyframes? 
	\item {\em Metric.} What metric should be used to give distance between a pair of frames in the feature space?
	\item {\em Matching.} How are the frames paired between the two summaries? 
\end{enumerate} 
The fourth question pertains to the form of $\gamma$ itself:
\begin{enumerate}
 \setcounter{enumi}{3}
 \item {\em Similarity.} Given a number of pairings between two keyframes sets, and the sizes of the two sets, what value do we assign to the similarity $\gamma$ of the sets?
\end{enumerate}

To address point 1), we examine the most widely used sets of features for representing keyframes. Typically, these features are colour-based (e.g., histograms of the hue value), summarising colour values for either the whole image or a grid-like split of the image into 2-by-2 and 3-by-3 subimages. We include in the comparison the RGB and HSV spaces, and other standard, though less popular, colour space representations.
We also take the last fully-connected layer of the ConvNet\footnote{\url{http://www.vlfeat.org/matconvnet/}}(Convolutional Network)~\cite{vedaldi15} pretrained model\footnote{Visual Geometry Group-Very Deep (VGG-VD-16): \url{http://www.robots.ox.ac.uk/~vgg/research/very_deep/}}~\cite{simonyan2014very} as a fixed feature extractor for our data, as well as SURF features.

For point 2), we consider the Euclidean and Manhattan distances for the feature spaces apart from SURF. For the SURF representation, we apply the associated method for matching relevant points between two images~\cite{Bay2008}.

For point 3), we describe and evaluate six approaches taken from the literature on keyframe evaluation. We believe that this is the first study which summarises and evaluates together these approaches. 

Finally, (for point 4) we propose the F-measure as $\gamma$ because of its symmetry, limits, and interpretability.

To determine the empirical answers of questions 1-3, we carry out an experimental study on the 50 videos from the VSUMM project~\cite{DeAvila2011} together with the automatic keyframe summaries and user ground-truth summaries provided. 


\begin{figure*}
	\centering
	\includegraphics[width=0.85\linewidth]{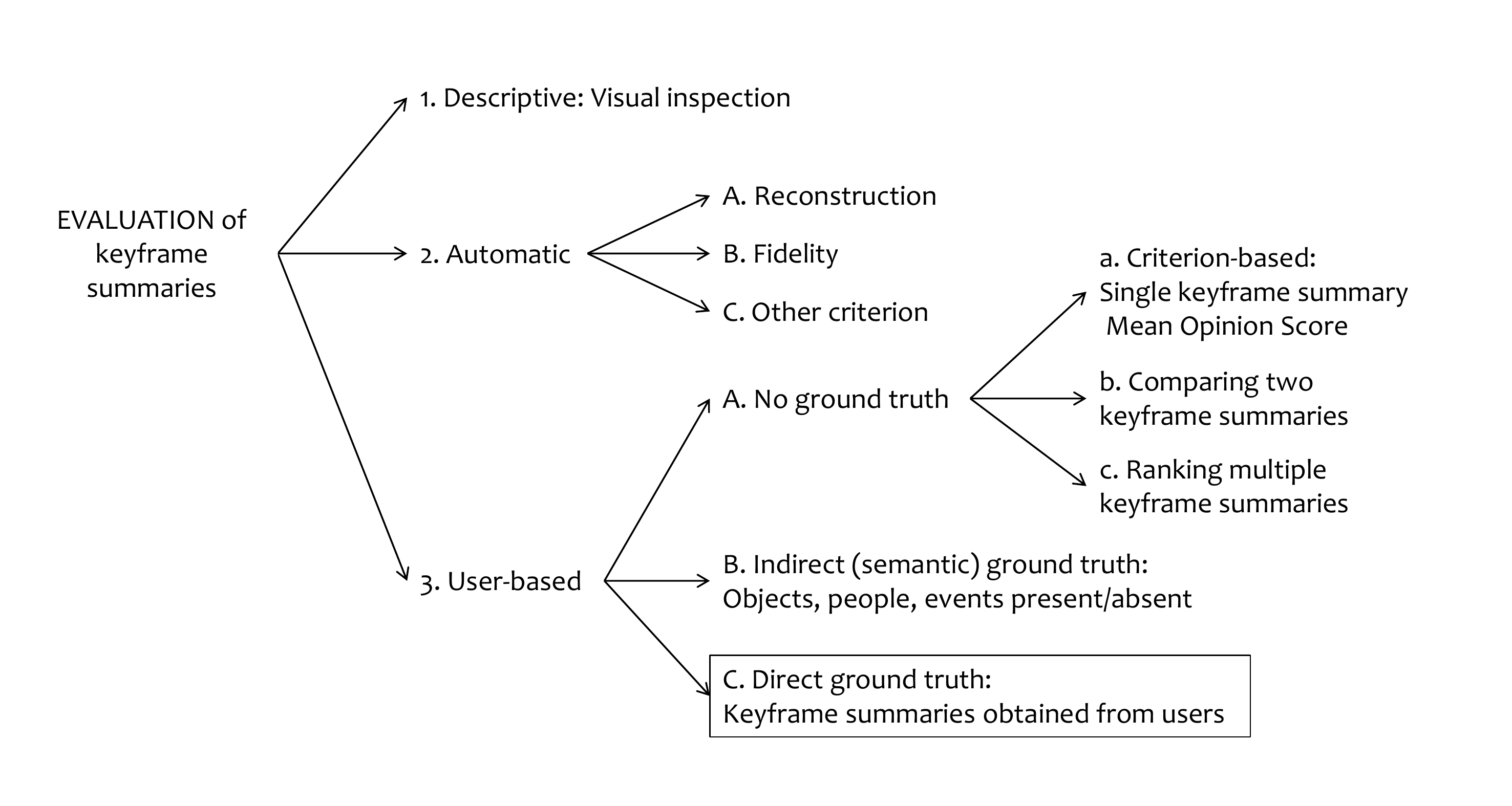}
	\caption{Evaluation approaches for keyframe video summaries. The box indicates the approach giving rise to the protocol proposed here.}
	\label{fig:EvaluationApproaches}
\end{figure*}

The rest of the paper is organised as follows. Section~\ref{sec:evo} gives a broad overview of existing evaluation practices, including those unrelated to ground-truth methods.  Sections~\ref{sec:fea} and \ref{sec:simframes} describe related work, and our experimental choices, in regard to questions 1) and 2) above, respectively.  Section \ref{sec:simset} gives the equivalent discussion of existing work and experimental choices for questions 3) and 4). 
Sections~\ref{sec:prop} and \ref{sec:exp} give the proposed protocol and its experimental evaluation using the VSUMM video repository, as well as a discussion on the findings. Section~\ref{sec:con} outlines our conclusions and further research directions. 

\section{Evaluation approaches}
\label{sec:evo}

A diagram summarising the most used evaluation approaches is shown in Figure~\ref{fig:EvaluationApproaches}. 

Category 1, ``Descriptive'' evaluation, pertains to some of the earliest publications in this area, which made no formal evaluation of their keyframe summaries, but simply displayed example outputs and argued for their plausibility (e.g. \cite{Sun2000}, \cite{Vermaak2002}, \cite{Yu2004}).  The field has since developed, giving rise to a great number of alternative summarisation algorithms as well as tools for quantitatively comparing their outputs. In the light of these developments, this ``descriptive'' (non-)evaluation must be considered an obsolete practice.

All forms of quantitative evaluation which involve no user input are grouped together under category 2.  Subcategory 2A includes those evaluation approaches, such as the Shot Reconstruction Degree~\cite{Liu2004}, which evaluate the keyframe selection by interpolating a reconstruction of the original video from it, and comparing this reconstruction to the original video.  To evaluate a keyframe selection in this way is to evaluate it, essentially, as a form of compression. However, there is no guarantee that the frame set which gives rise to the best reconstruction is the one which a human user would choose.  Indeed, this is the very reason that summarisation is considered as a distinct task from compression.

Subcategory 2B includes the Fidelity measure of Chang et al. \cite{Chang1999}.  This involves using a semi-Hausdorff measure to calculate a distance from the keyframe set to the set of frames of the original video. The Hausdorff and semi-Hausdorff distances are generic means of calculating a degree of similarity or distance between two sets. They are mathematically convenient and familiar from their applications in other fields, but may be inappropriate for the problem of evaluating keyframe sets, due to their strong sensitivity to the distance of the worst-case (most distant) element.  

Subcategory 2C characterises methods which use a high-level feature regarded as measuring ``quality'' to gauge the keyframe summary. Often this feature is used itself to select the keyframes, so there is no active comparison process with the whole video. Examples from this category include snapshot detection~\cite{Xiong2014} where the main objective of the keyframe selection is to be like a well-composed photograph taken by a human. 

All forms of evaluation which involve user input in one way or another are grouped together under category 3.  Subcategory 3A contains those methods in which users are asked to assess the automatic keyframe summary.  Subcategories 3B and 3C contain those methods in which users generate a summary of their own, which is then used as a ground-truth against which the output of the algorithm is automatically compared.

Following Truong and Venkatesh \cite{Truong2007}, we distinguish between a ``direct'' ground truth, in which the user makes a keyframe summary, and an ``indirect'' ground truth, in which the user summarises the semantic content which the output keyframe summary should cover.  The former subcategory, 3C, includes the keyframe-matching method developed by de Avila et al.~\cite{DeAvila2011}, which has gained some popularity (e.g. \cite{Ejaz2012}, \cite{Gong2014}, \cite{Mei2015}).

The bulk of current interest seems to be in evaluations of the type covered by category 3. In this paper we develop further the approach 3C. 
Given that ground-truth summaries for standard datasets are now publicly available, it makes sense to hone keyframe selection methods using this shared standard data. 
Subject to an accepted protocol, the use of established ground truths has the great advantage of providing a method for comparing keyframe summaries with one another, and with baseline methods. Such a protocol is objective, unified, and inexpensive.

\section{Features}
\label{sec:fea}

The feature spaces used for {\em evaluating} keyframe summaries are typically quite different from the feature spaces used by the various algorithms for selecting the keyframes for their summaries.  Selection methods often rely on sophisticated and context-involved features such as 
the presence of people,  objects ~\cite{Lu2013,Chao2010,Lee2015,Lee2012}, landscapes, motion~\cite{Liu2003,Ejaz2012,Varini2015}, or famous landmarks~\cite{Gygli2014}, and/or the use of a visual thesaurus~\cite{Spyrou2009}. High-level descriptors gauging the quality of the video frames have also been proposed, for example, ``aesthetics'', ``attention'', ``saliency'' and ``interestingness''~\cite{Gygli2014}.

For judging the similarity between a user keyframe collection and a candidate keyframe collection, however, low-level, context-blind features are usually applied. Feature spaces of this type include colour histograms, and edge  and texture features~\cite{Doherty2008,Wang2012,Cooper2005,Vermaak2002,LakshmiPriya2014,Sun2000,DeAvila2011,Ohta1980,Lin2006}.

In this study we look for a suitable feature representation of the keyframes among the alternatives listed below. Most of the feature sets are defined by splitting the image into 3-by-3 equal-sized subimages before extracting features from each sub-image: this is the meaning of the \textit{\_9blocks} suffix.
\begin{enumerate} 	
 	\item {\em RGB\_9blocks}. The mean and standard deviations of the red ($R$), green ($G$) and blue ($B$) channels for each subimage. (6 features per sub-image)
 	\item {\em HSV\_9blocks}. The mean and standard deviations of HSV (6 features per sub-image)~\cite{DeAvila2011,Zhuang1998}.
 	\item  {\em CHR\_9blocks}. The mean and standard deviations of Chrominance components $C_1$ and $C_2$ (4 features per sub-image) calculated as~\cite{Vermaak2002}:
\[
C_1 = \frac{R}{q},\;\;  C_2 = \frac{G}{q},\;\;q=\sqrt{R^2+G^2+B^2},
\]
 	\item  {\em OHT\_9blocks}. The mean and standard deviations of features $I_1$, $I_2'$ and $I_3'$ of Ohta space (6 features per sub-image) calculated as~\cite{Ohta1980}
\begin{eqnarray*}
I1 &=& \frac{1}{3}(R + G + B)\nonumber\\
I2' &=& R-B\nonumber\\
I3' &=& \frac{1}{2}(2G - R - B) \nonumber
\end{eqnarray*}
\end{enumerate} 	
The following hue histogram feature spaces (H-histograms) were also investigated~\cite{Gong2003, Lin2006, Wang2012, DeAvila2011, Hanjalic1999, Zhuang1998, Uchihashi1999, Furini2010, Zhu2004}:
\begin{enumerate}
  \setcounter{enumi}{4}
	\item {\em H8\_9blocks}. A histogram of the hue (H) values of the HSV space with 8 bins (8 features) for each of the sub-images of a split of the image into a 3-by-3 grid.
 	\item {\em H16\_1block}. H-histogram with 16 bins for the whole image. 
 	\item {\em H16\_4blocks}. H-histogram with 16 bins for a 2-by-2 split of the image in sub-images.
 	\item {\em H16\_9blocks}. H-histogram with 16 bins for a 3-by-3 split of the image in sub-images.
  	\item {\em H32\_1block}. H-histogram with 32 bins for the whole image. 
\end{enumerate}
The values of each histogram were scaled so that the sum was one.
 
Next we considered: 
\begin{enumerate}
\setcounter{enumi}{9}
\item {\em CNN. } The last fully connected layer of a pre-trained CNN was used as a 4096-dimensional feature space~\cite{simonyan2014very}.

\item{\em SURF.} SURF features were extracted which could match relevant points between two images~\cite{Apostolidis2014,Ratsamee2015,Jinda2013}.
\end{enumerate}





\section{Similarity between two keyframes}
\label{sec:simframes}

Similarity between two images (keyframes) can be calculated in many ways. For example, one could evaluate the proportion of matched SIFT keypoints~\cite{Liu2009}, or similarities between visual word histograms~\cite{li2010vert,Spyrou2009}. However, more general and efficient similarity measures can be used if the images are represented as points in an $n$-dimensional feature space.  This will be our approach with the first 10 of the feature sets specified above.  The SURF approach will be our example of an alternative approach that does not attempt to embed the keyframes in $\mathbb{R}^n$.


Here we treat the collections of keyframes as unordered. We use the Manhattan distance\footnote{The Manhattan distance is the Minkowksi distance with $p=1$, i.e. the L1 norm.} and the Euclidean distance on each of the feature spaces 1--10. 

For the SURF features, we use the following procedure: 1. Identify the keypoints in frame 1 (total number $n_1$), and find how many have been matched in frame 2 (say, $k_1$). 2. Identify the keypoints in frame 2 ($n_2$) and the number of matched keypoints in frame 1 ($k_2$). Calculate the similarity between the two images as the proportion $\frac{2\min\{k_1,k_2\}}{n_1+n_2}$. For the sake of consistency, we will use instead a distance between two frames $f_a$ and $f_b$ calculated as 
\[
d(f_a,f_b) = 1 - \frac{2\min\{k_1,k_2\}}{n_1+n_2}.
\]

\section{Similarity between two sets of keyframes}
\label{sec:simset}

To evaluate an automatic keyframe summary, we can compare its match to a ground-truth summary by counting the number of paired frames and taking into consideration the total number of frames in each summary~\cite{DeAvila2011,Ejaz2012,Gong2014}, \cite{Mei2015}. 

We assume that a distance measure $d$ between two frames has been already chosen, as discussed in section \ref{sec:simframes}. Two frames $f_a$ and $f_b$ are sufficiently similar to be called a match if $d(f_a,f_b)<\theta$, where $\theta$ is a chosen threshold.

Let $K_1$ and $K_2$ be two sets of keyframes. We are interested in a measure of closeness between the two sets, $M(K_1,K_2)$. The following two questions must be answered: How do we count the number of matches $m$ between $K_1$ and $K_2$? Once $m$ has been found, how do we use it to calculate $\gamma$? (These are questions 3) and 4) of the introduction.)

\subsection{Finding the number of matches}
\label{sec:simset1}
Denote the cardinalities of the two summaries by $N_1=|K_1|$ and $N_2=|K_2|$. Construct a distance matrix $D_{(N_1\times N_2)}$ where entry $d_{i,j}$ in $D$ is the distance between frames $i\in K_1$ and $j\in K_2$. Denote the number of matches returned by $m$. Apart from the Mahmoud algorithm below, all algorithms take as input $D$ and $\theta$, and return $m$.

Here we examine six pairing (matching) algorithms:
\begin{enumerate}
\item Na\"ive Matching (no elimination).  This algorithm is surprisingly popular \cite{Mahmoud2014, aherne1998bhattacharyya} although it has an obvious flaw. If the candidate summary $K_1$ consists of nearly identical frames which happen to be close to one frame from the ground truth summary $K_2$, then the number of matches will be perfect, $m=N_1$, for an arbitrary $N_1$. Such a candidate summary, however will be quite inadequate: it is neither concise nor representative. Algorithm~\ref{NaiveMatching} relies on the presumption that $K_1$ is a reasonable summary containing diverse frames.
\item Greedy Matching. This algorithm is widely used but is quite conservative.
\item Hungarian Matching.~\cite{Khosla2013} The Hungarian algorithm will identify $\min\{N_1,N_2\}$ pairs such that the sum of the distances of the paired frames is minimum. A thresholded matching can be na\"ively formed from this minimal complete matching by simply removing all pairings over the threshold distance $\theta$. Thus, close matches could be missed in an attempt to minimise the total distance.
\item Mahmoud algorithm.~\cite{Mahmoud2014} For this algorithm, the frames are arranged in temporal order and the matches are checked and eliminated accordingly. Apart from the temporal ordering, the algorithm is identical to the Greedy Matching. 

\item Kannappan algorithm.~\cite{Kannappan16} An interesting alternative approach to the matching problem is put forward by Kannappan et al.~\cite{Kannappan16}. In their approach, a keyframe from the candidate set and  a keyframe from the ground truth are matched only if each is the other's best possible match: Algorithm~\ref{Kannappan}.  In their implementation, the set of matched pairs is subsequently thresholded using a different concept of pairwise frame distance from that used to form the matches.  We have modified this procedure to make it the equivalent of the de Avila et al.\ thresholding, by using the same distance metric for thresholding as for finding the pairings.

\item Maximal Matching. The greatest possible value of $m$ is given by a maximal \emph{unweighted} matching in which frames less than distance $\theta$ apart can be paired. Such a matching is given by the Hopcroft-Karp algorithm~\cite{West}. We will use instead the convenient alternative Algorithm~\ref{NotBloodyWellHopcroft}, in which we find the lowest-weight complete matching on a binary matrix $D'$ obtained by thresholding $D$. Entry $d'_{i,j}$ in $D'$ has value 0 if $d_{i,j}<\theta$, and 1 otherwise. After the optimal assignment is found through the Hungarian algorithm, the number of matches is determined by counting how many of the matched pairs are at distance less than $\theta$.
\end{enumerate}

\begin{algorithm}
	\DontPrintSemicolon
	$m\gets 0.$

	\For {$i=1,\ldots,N_1$}
	{If any $d_{i,j}< \theta$, $j=1,\ldots,N_2$, increment the number of matches, $m \gets m+1$.}

	\caption{Na\"ive Matching}
	\label{NaiveMatching}
\end{algorithm}

\begin{algorithm}
	\DontPrintSemicolon
	$m\gets 0.$
	Find the smallest distance $d_{\min} = \min D$. 

	\While {$d_{\min}< \theta$} 
	{Increment the number of matches, $m \gets m+1$.
	
	Remove the row and the column of the matched elements from $D$.
	
	Find the smallest distance from the remaining matrix $d_{\min} = \min D$.} 

	\caption{Greedy Matching}
	\label{GreedyMatching}
\end{algorithm}

\begin{algorithm}
	\DontPrintSemicolon
    Apply the Hungarian assignment algorithm to $D$.

    Identify the matched pairs of frames $(i,j)$, and retrieve the distances $d_{i,j}$ from $D$.

    Assign to $m$ the number of these distances which are smaller than $\theta$.

	\caption{Hungarian Matching}
	\label{Hungarian}
\end{algorithm}

\begin{algorithm}
	\DontPrintSemicolon
	
	\medskip
	\KwIn{Keyframe summaries $K_1$ and $K_2$ arranged in temporal order, and threshold $\theta$.}
	
	\medskip
	\KwOut{Number of matches $m$.}

	\medskip
	$m\gets 0.$

	\For {$i \in K_1$} 
	{\For {$j \in K_2$}
	{\If {$d_{i,j}<\theta$,}
		{Increment the number of pairings, $m \gets m+1$.
	
		Remove $i$ from $K_1$ and $j$ from $K_2$.

		Break.}
	}
    }
	\caption{Algorithm of Mahmoud \cite{Mahmoud2014}}
	\label{MahmoudMatching}
\end{algorithm}

\begin{algorithm}
	\DontPrintSemicolon

	\medskip
	Initialise a set of pairings $M\gets \emptyset.$

	\For {each frame  $i \in K_1$} 
	{\For {each frame $j \in K_2$}
	{\If {$j' = \arg \min_{k\in K_2} d(i,k)$ and
		$i' = \arg \min_{k\in K_1} d(k,j)$}
		{Add the pair to the matching set $M\gets M\cup\{(i',j')\}$.}
	}
	}
	Remove from $M$ all pairs for which $d(i',j') \geq \theta$.

   $m\gets |M|.$

	\caption{Algorithm of Kannappan et al.~\cite{Kannappan16}}
	\label{Kannappan}
\end{algorithm}

\begin{algorithm}
	\DontPrintSemicolon

    Construct matrix $D'$ of the same size as $D$ such that $d'_{i,j} = 0$ iff $d_{i,j} < \theta$, and  $d'_{i,j} = 1$, otherwise.

    Apply the Hungarian assignment algorithm to $D'$.

    Identify the matched pairs of frames $(i,j)$, and retrieve the distances $d_{i,j}$ from $D$.

    Assign to $m$ the number of these distances which are smaller than $\theta$.

	\caption{Maximal matching algorithm}
	\label{NotBloodyWellHopcroft}
\end{algorithm}

A common drawback of these algorithms is the lack of guidance in choosing the threshold value $\theta$. This value has an immediate impact on the number of matches, and subsequently on the value of the measure $\gamma(K_1,K_2)$. Different values may be appropriate for different feature spaces and metrics. While the L1 distance between distributions, such as elements of histogram feature spaces, is bounded between 0 and 2, the same is not true for other feature spaces. For histogram spaces, $\theta = 0.5$ has been empirically found useful~\cite{DeAvila2011} but it is not clear what theoretical meaning this value might have. Setting an interpretable threshold in other feature spaces is even less intuitive. 

For our experiments, we will use a range of thresholds from 0.01 up to 0.7 for the Manhattan metric.  For the Euclidean metric, we will scale the threshold relative to the distribution of all pairwise distances between frames in the video. The thresholds will be percentiles of this distribution, from the 0.01th up to the 3rd percentile. For the SURF metric, we will vary the threshold between 0.01 and 0.4.
 
\subsection{Calculating the similarity between keyframe summaries using the number of matches}
\label{sec:simset2}
Interpreting the number of pairings $m$ returned by their Greedy Matching algorithm, de Avila et al.~\cite{DeAvila2011} use a pair of measures called respectively ``Accuracy rate'' ($CUS_A$) and `` Error rate'' ($CUS_E$), both designed to express how well $K_1$ (candidate summary) matches $K_2$ (ground truth), but not the other way around: 
\[
CUS_A = \frac{m}{|K_2|},
\]
where $|\zeta|$ denotes the cardinality of set $\zeta$, and 
\[
CUS_E = \frac{|K_1|-m}{|K_2|}.
\] 
The problem with these measures is that the upper limit of $CUS_E$ depends on $|K_1|$. 

Alternatively, given a number of matches $m$, the similarity between $K_1$ and $K_2$ can be quantified using the F-measure, whose advantage is that it is symmetric on its two arguments~\cite{Gong2014}. Without loss of generality, choose $K_1$ for calculating the Recall, and $K_2$ for calculating the Precision. Then
\begin{eqnarray}
{\rm Recall} &=&\frac{m}{|K_1|}\nonumber\\ 
{\rm Precision} &=&\frac{m}{|K_2|}\nonumber\\
F(K_1,K_2) &=& \frac{2({\rm Recall}\times {\rm Precision})}
{{\rm Recall} + {\rm Precision}}\nonumber\\
&=& \frac{2m}{|K_1|+|K_2|}
\label{f}
\end{eqnarray}
We have chosen to use this $F$-measure as our $\gamma(K_1,K_2)$ because, unlike $CUS_A$ and $CUS_E$, it is symmetric, limited between 0 and 1, and interpretable.

We note that there is a potential problem when using the $F$-measure with the Naive Matching algorithm and the Kannappan algorithm because they do not guard against $m>N_2$, which may lead to $F>1$. In such cases we clipped the value of $F$ to 1.

\section{Proposed evaluation protocol}
\label{sec:prop}

We have reviewed the approaches and methods to answer the four questions in the Introduction: (1) Features in Section~\ref{sec:fea}; (2) Metric in Section~\ref{sec:simframes}; (3) Matching in Section~\ref{sec:simset1} and (4) Similarity Measure in Section~\ref{sec:simset2}.
 
The foundational idea for our experiments is that a good measure for similarity between keyframe summaries should distinguish as clearly as possible between content-blind baseline methods such uniform summaries on the one hand, and a sophisticated algorithmic summary, on the other hand.  To estimate how well a measure distinguishes between baseline designs and bespoke selection methods, we propose the quantity which we call ``discrimination capacity'' as the difference:
\begin{equation}
c_U \overset{\Delta}{=}  c_U(K,U,GT) = \gamma(K,GT)-\gamma(U,GT),
\label{eq:dc}
\end{equation}
where $GT$ is a ground truth summary, $K$ is a keyframe summary obtained by an algorithmic method, and $U$ is a baseline summary, which in our case will be the Uniform summary of the same cardinality as $K$. Large values of  $c_U$ will signify good choices of parameters $\Theta$: features, metrics, algorithms, and thresholds which could be recommended for the practical implementation of the proposed protocol as a tool for the evaluation of future algorithms. 

For the sake of generality, our protocol is bound by minimal restrictions:
\begin{enumerate}
	\item We are not concerned with {\em how} the keyframes in $K$ are obtained. For example, the video could be split into shots or used in its entirety; low-level visual features or high-level semantic features could be used, etc.
	\item Both $K$ and the ground truth summaries are {\em sets} of keyframes. This means that the frames are not ranked by importance, nor are they arranged in a temporal order.
\end{enumerate}
Weighing the arguments for and against fusing a possible set of available ground truth summaries into a single summary~\cite{huang2004automatic,Gong2014}, we decided not to include a fusing procedure, in order to maintain simplicity and transferability. Such a procedure could be designed in many different ways, and there is little to guide the choice. We opt for calculating the overall assessment of $K$ as the average of the measures of interest between $K$ and the ground truth summaries. For example, let $G=\{K_{g1},\ldots,K_{gL}\}$ be a collection of ground truth summaries obtained from $L$ users. Let $U(k)$ be a uniform summary with $k$ keyframes. We calculate $C_U$, the average of $c_U$ for $K$ and $G$, as
\begin{eqnarray}
C_U &=&\frac{1}{L}\sum_{i=1}^L c_U(K,U,K_{gi})\nonumber\\
&=& \frac{1}{L}\sum_{i=1}^L \left( F(K,K_{gi}) - F(U(|K|),K_{gi}) \right) \;. 
\label{cu}
\end{eqnarray}
This value measures how much better $K$ is, compared to a uniform keyframe summary of the same cardinality, in matching the users' views. Note that, for a given $K$ and $G$, $C_U$ depends on the choices we make for the parameters in $\Theta$: features, metric, pairing algorithm and threshold. Therefore we will be looking for a set of parameters which maximises $C_U$ across a range of videos and summarisation algorithms for obtaining $K$.

\section{Experimental study}
\label{sec:exp}

\subsection{Data and set-up}
For this experiment we used the VSUMM collection\footnote{ \url{https://sites.google.com/site/vsummsite/download}}, containing 50 coloured videos in MPEG-1 format (30 fps, 352 *240 pixels). Videos cover several genres (e.g. documentary, educational, historical) with various duration from 1 to 4 minutes.  Each video has been manually summarised by 5 different users. 

The purpose of the experiment is to identify a set of choices of feature space, metric, algorithm, and threshold, which maximises the discrimination capacity $C_U$~(\ref{cu}). 

We considered: 11 feature spaces, 6 matching algorithms, 2 concepts of distance (Euclidean and Manhattan) for the metric spaces and a proportion-based distance for the SURF features, and a range of values of the threshold $\theta$ for each distance. 

For the Uniform baseline, for each video we generated 30 summaries with cardinalities from 1 to 30.  To generate a summary with $k$ frames, the video was split into $k$ consecutive segments of approximately equal length, and the middle frame of each segment was taken in the summary.


\begin{figure}[htb]
	\centering
	{\includegraphics[width=\linewidth]{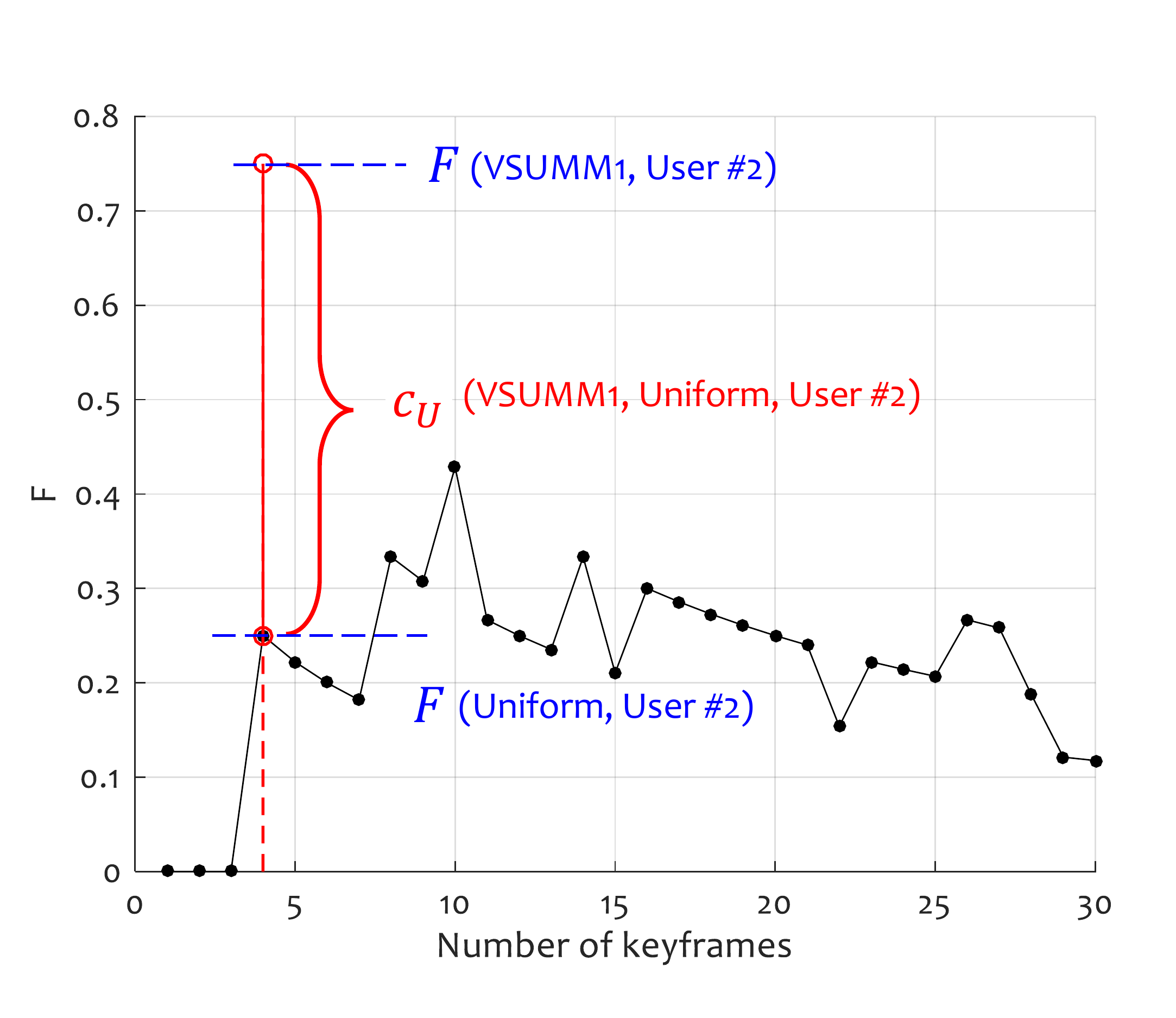}}
	\caption{An example of calculating $c_U$ for the VSUMM1 keyframe selection method, video \#22, feature space \#6 (H16\_1block), the Hungarian Matching method, Manhattan distance and threshold 0.5. $c_U$ is the difference between the F value for matching candidate summary VSUMM1 to User \#2 (ground truth \#2) and the F value matching a uniform summary of the same cardinality as VSUMM1 (4 in this case) and User \#2. $C_U$ is the average of the 5 such $c_U$ terms in eq.(\ref{cu}).} 
	\label{fig:CUCR}
\end{figure}

Figure~\ref{fig:CUCR} illustrates graphically the calculation of one term of $C_U$ in eq.(\ref{cu}). We chose to pair the number of uniform keyframes with the number of keyframes in the summary of interest in order to make a fair comparison. The value of $C_U$ is a measure of ``how much closer the summary is to a ground truth compared with a uniform summary of the same size''. Naturally, we will be looking for a combination of parameters $\Theta$ which maximises $C_U$ across the video collection in this experiment.

For the full calculation of $C_U$ for this example, we need the remaining four terms as shown in Table~\ref{tab:example}.

\begin{table*}[htb]
	\caption{An example of calculation of $C_U$ for  the VSUMM1 keyframe selection method, video \#22, feature space \#6 (H16\_1block), the Hungarian Matching method, Manhattan distance, and threshold 0.5. The $F$-values are shown in the table; the bottom row contains the terms in (\ref{cu}); the values for user \#2, marked with * are the ones in Figure~\ref{fig:CUCR}.}
	\label{tab:example}
\centering
\begin{tabular}{rcccccc}
User \# &1 &    2* &    3 &   4 &   5& $C_U$\\
\hline
$F(VSUMM1,user)$&    0.5000&    0.7500*&    0.6667&    0.2857&    0.4444&\\
$F(U_4,user)$&   
0.5000&    0.2500*&    0.2222&    0.2857&    0.4444&\\
\hline
Term&    0&    0.5000*&    0.4444&         0&         0&
    {\bf 0.1889}\\
\end{tabular}
\end{table*}

In our experiments we calculated $C_U$ for every choice of parameter settings and every video. The algorithmic summarisation methods used are the 5 methods provided within the VSUMM video data base: Delaunay Triangulation (DT) ~\cite{Mundur2006}, Open Video Project (OV)\footnote{\url{https://www.open-video.org.}}, STIll and MOving Video Storyboard (STIMO)~\cite{Furini2010}, Video SUMMarization1 (VSUMM1)~\cite{DeAvila2011}, and Video SUMMarization2 (VSUMM2)~\cite{DeAvila2011}.

\subsection{Evaluation of distance metric and threshold for similarity between frames}
\begin{figure*}[htb]
	\centering
\begin{tabular}{ccc}
	\includegraphics[width=0.31\linewidth]{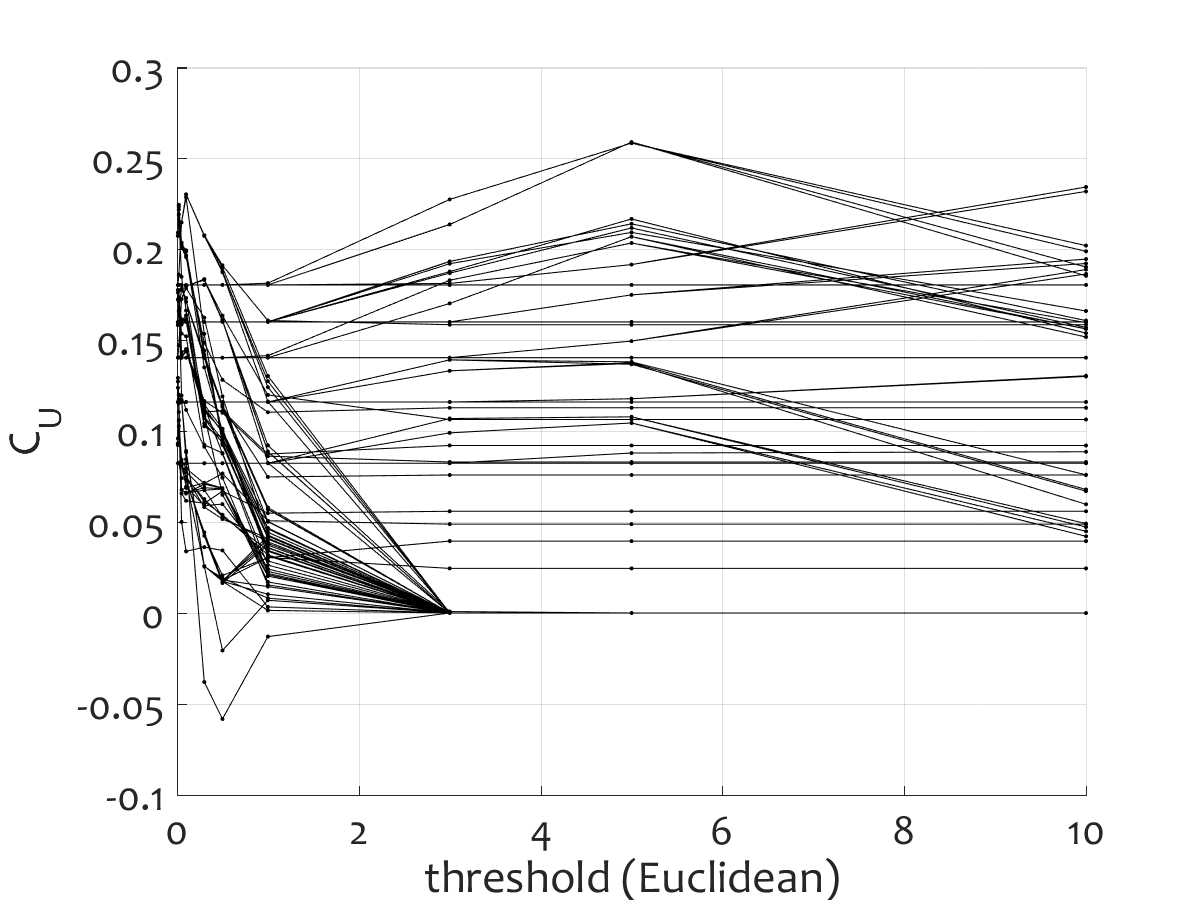}&
    \includegraphics[width=0.31\linewidth]{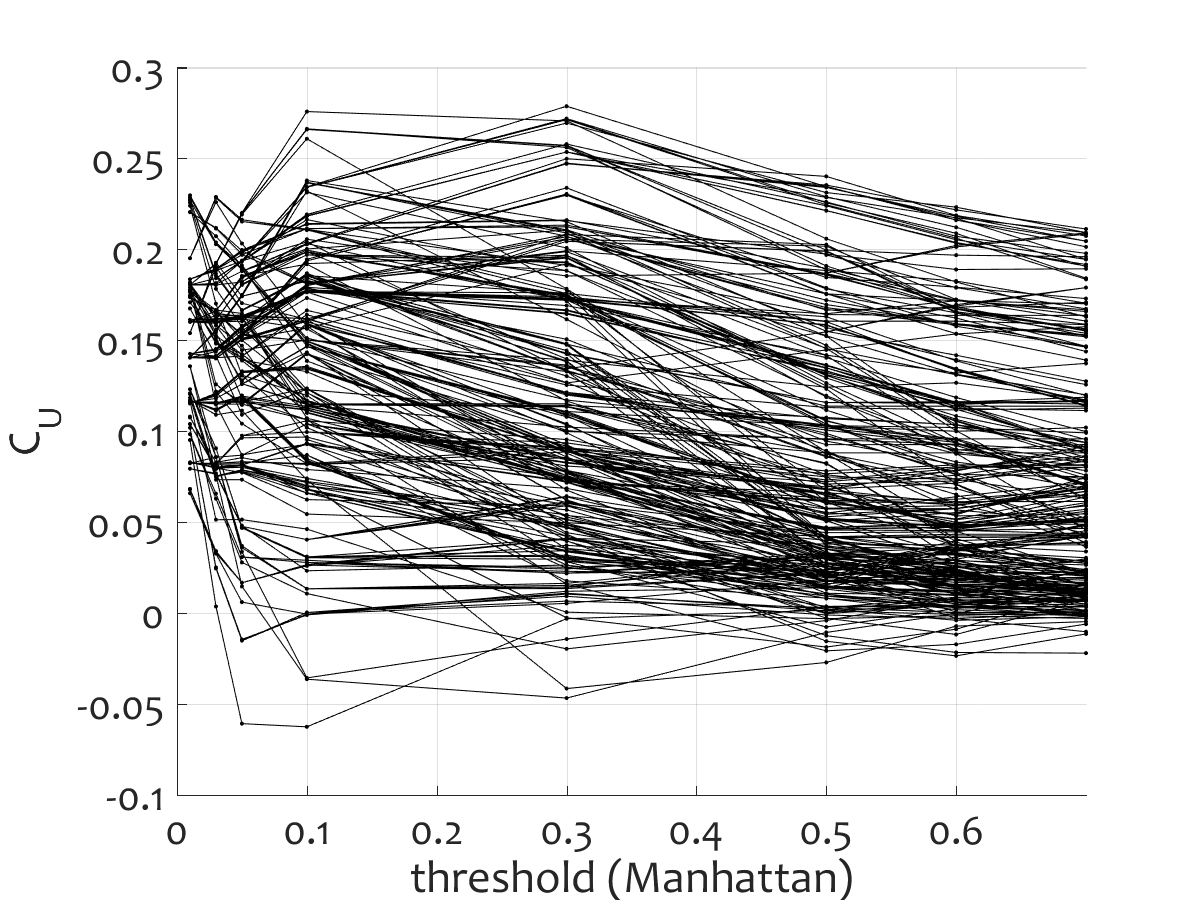}&
	\includegraphics[width=0.31\linewidth]{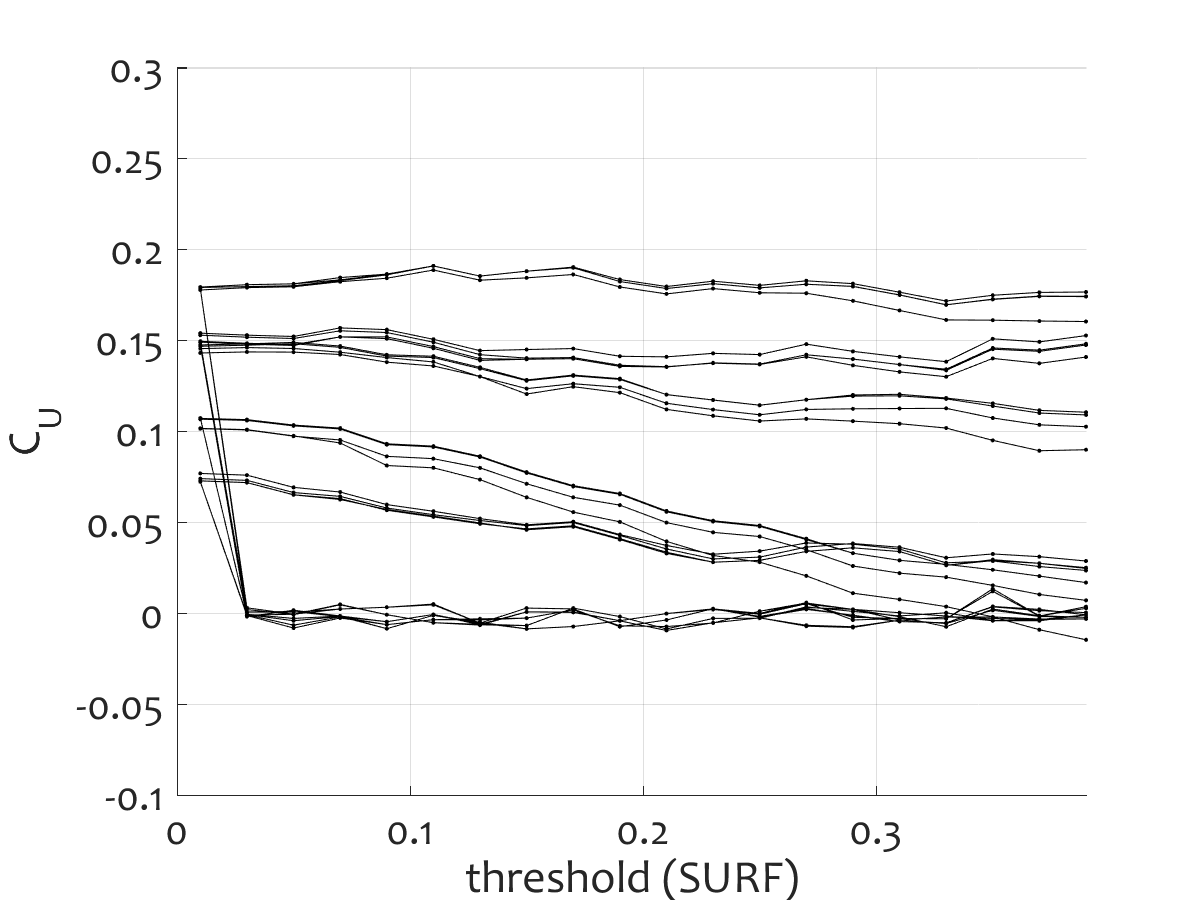}\\
&&\\
(a) Euclidean distance&
(b) Manhattan distance& 
SURF feature distance\\
\end{tabular}
\caption{Discrimination capacity $C_U$ as a function of the threshold for the three types of distances used. Each of plots (a) and (b) contains 300 line graphs (10 feature spaces, 6 matching methods, 5 summarisation methods). Plot (c) contains 30 lines (SURF space, 6 matching methods, 5 summarisation methods. Each line is the average across 50 videos and 5 users.} 
\label{fig:CUall}
\end{figure*}

As the choice of threshold ranges were only guessed to be suitable, comparing averages across all threshold values may be misleading. Therefore we plot $C_U$ for {\em all} the feature spaces, matching methods, and summarisation methods as a function of the threshold. Figure~\ref{fig:CUall} shows these plots. Note that $C_U$ may be negative. This is the undesirable case where the uniform summary matches the user ground truth better than the algorithmic (candidate) summary.

The shape of the line graph in relation to the threshold is expected to be convex with lower values for smaller and larger thresholds. For small thresholds, there will be very few matches, hence the F-values will be low for both the candidate summary and the uniform summary, hence the difference $C_U$ will be small. For large values of the threshold, a large number of matches will be detected in both comparisons, both F-values will be high, and the difference $C_U$ will be small again.  The best results (larger $C_U$) are offered by the Manhattan distance. The peak for the Manhattan distance is between $\theta = 0.3$ and $\theta = 0.5$. For the Euclidean distance, there are two different types of curves. Some peak quite early, at $\theta$ between 0 and 0.5, while others stay stable. The SURF feature curves exhibit consistent and stable patterns which will be analysed later. From these findings, we favour the Manhattan distance for our proposed protocol, and will use this distance for the following evaluation of the feature spaces.

\subsection{Evaluation of feature spaces}
\begin{figure}[htb]
	\centering
	\includegraphics[scale = 0.85]{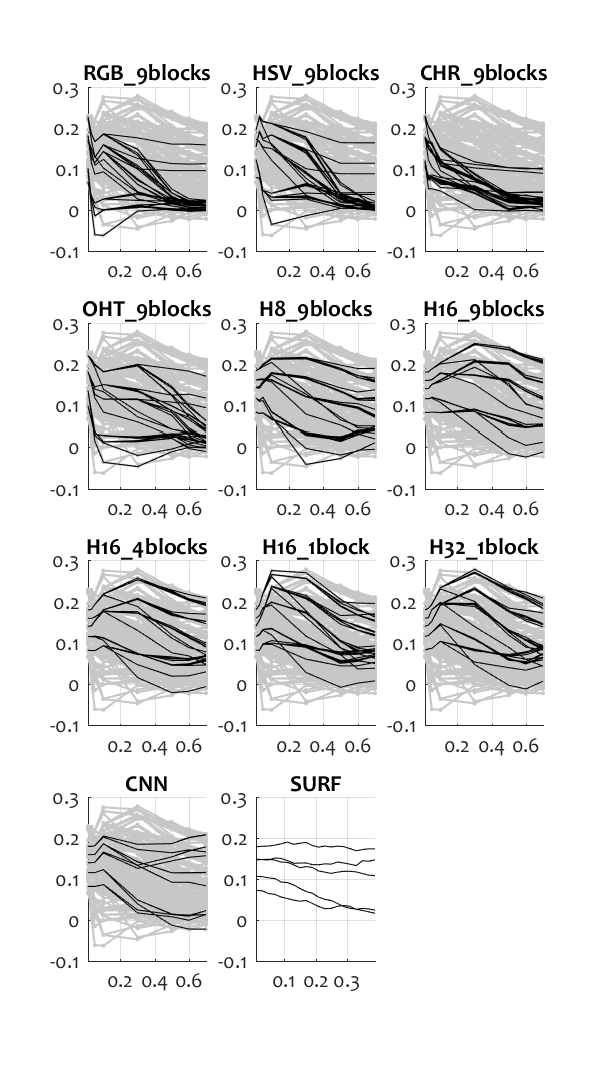}
\caption{Discrimination capacity $C_U$ as a function of the threshold (Manhattan distance) with the 11 feature spaces.} 
\label{fig:fs}
\end{figure}

We look for a feature space which maximises the desirable quantity $C_U$. As the Manhattan distance gave the best results in the previous section, we will consider only this metric here. Figure~\ref{fig:fs} shows the results for the 10 feature spaces. Each sub-plot corresponds to one feature space. As in Figure~\ref{fig:CUall}~(b), the horizontal axis is the threshold used with the Manhattan distance, and the vertical axis is $C_U$. This time, all curves corresponding to the respective feature space are highlighted in black (30 such curves for each feature space: 6 matching methods, 5 summarisation methods).

Our results show that the simple colour spaces (1-4) are not useful in this context. The hue histograms, on the other hand, give the best results. The feature space with the largest $C_U$ is {\em H32\_1block}. This is somewhat surprising because the expected winner was either CNN or SURF, being high-level features. This result hints to the possibility that spending a lot of computational effort for calculating highly sophisticated properties of images may be unjustified in some cases. 
Thus, we propose to use {\em H32\_1block} for the purposes of automatic evaluation of keyframe summaries when ground truth is available.

\subsection{Evaluation of matching algorithms}
\begin{figure*}[htb]
	\centering
\begin{tabular}{ccc}
	\includegraphics[width=0.31\linewidth]{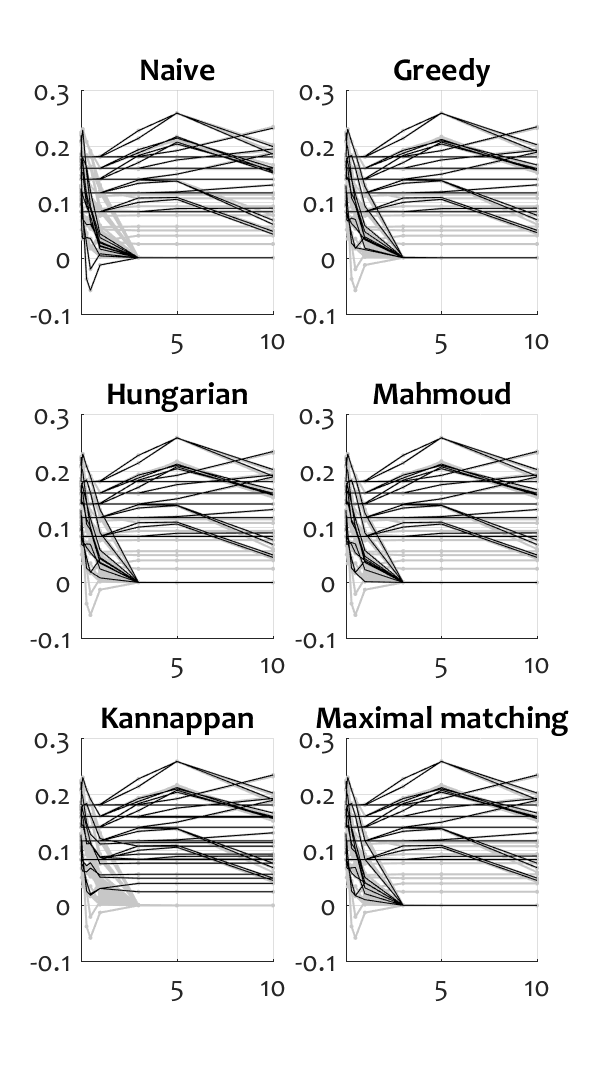}&
    \includegraphics[width=0.31\linewidth]{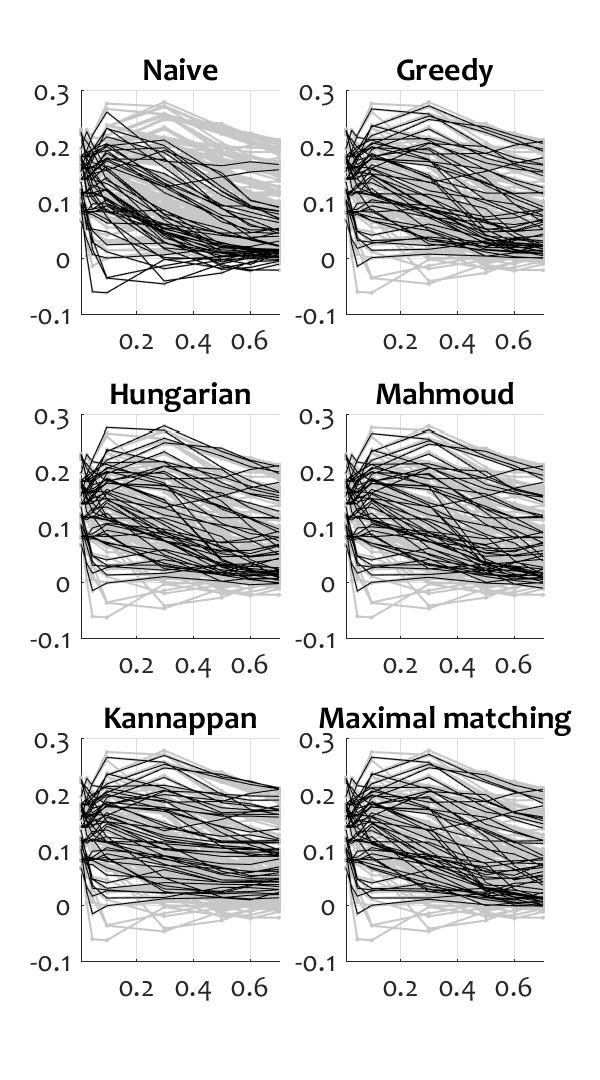}&
	\includegraphics[width=0.31\linewidth]{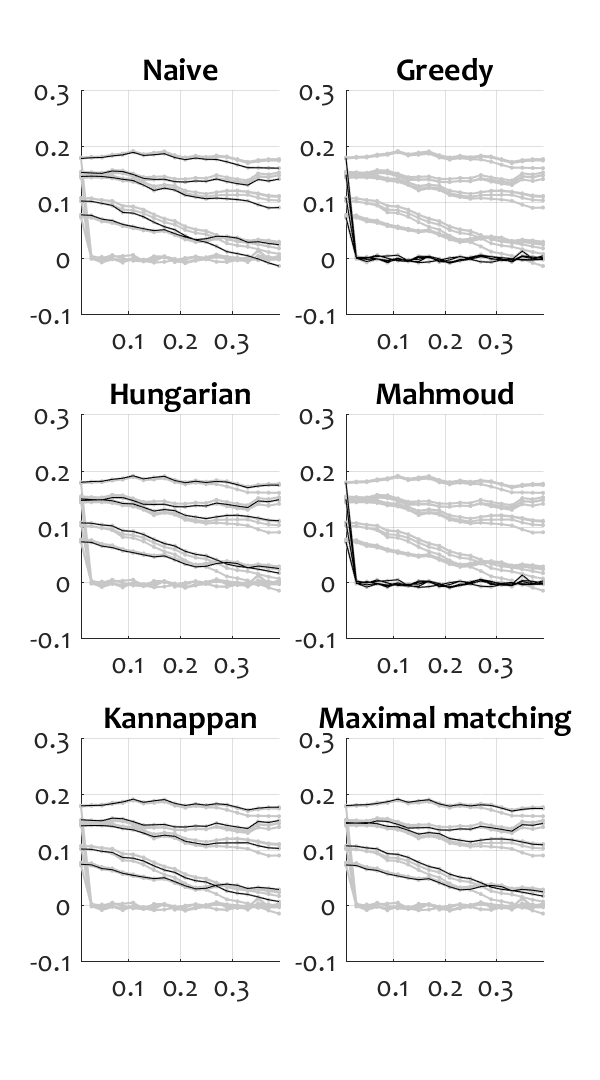}\\
(a) Euclidean distance&
(b) Manhattan distance& 
SURF feature distance\\
\end{tabular}
\caption{Visualisation of the $C_U$ for the 6 matching methods.} 
\label{fig:mm}
\end{figure*}

The results for this part are shown in Figure~\ref{fig:mm}. The format is the same as in Figure~\ref{fig:fs}. The lines plotted in black are the ones corresponding to the matching method in the title of the subplot.

It can be seen that, for Euclidean and Manhattan distance, the Naive matching is slightly inferior to the rest of the matching methods. This is to be expected, as the Naive labelling method may result is a large number of false positive matches for both the uniform summary and the summary of interest. This will smear the difference between the F-values, leading to low $C_U$. The remaining 5 methods are not substantially different. Interestingly, the conservative matching methods - Greedy and Mahmoud, do not work well with the SURF features. Note that here we view {\em all} the results together, both good and bad. Further analyses show that the variability in the $C_U$ for each matching method is not due to feature spaces but to summarisation method. The best such method, VSUMM1, corresponds to the highest curves.

Based on these results, we can recommend any of the three matching methods: Hungarian (minimal-weight complete matching followed by thresholding); Kannappan (The algorithm of Kannappan et al.~\cite{Kannappan16}); and Hopcroft-Karp (The Hopcroft-Karp algorithm or any equivalent algorithm returning a maximal unweighted matching from the sub-threshold pairings). Of these, Kannapan has the lowest computational complexity $O(n^2)$ compared with $O(n^3)$ for Hungarian, and with the maximal-matching method whose worst-case is $O(n^{2.5})$ if implemented as the Hopcroft-Karp algorithm, or $O(n^3)$ if implemented as algorithm~\ref{NotBloodyWellHopcroft}.  Hence we include the algorithm of Kannappan et al. in our proposed protocol.

\subsection{The proposed protocol, with example application}
\label{proposed}
Several authors (e.g. \cite{cahuina2013new,gong2014diverse,mei2015video}) have followed the choice of feature space, metric, algorithm, and threshold pioneered by de Avila et al.~\cite{DeAvila2011}. These choices seem to have had no previously published theoretical or experimental basis. The choice of H16\_1block feature space, and threshold value $\theta=0.5$ is reasonable, though the finer-grained H32\_1block feature space outperforms it on average.  

We propose the use of the following:
\begin{itemize} 
\item Feature space: 32-bin hue histogram  {\em H32\_1block} (normalised to sum 1), 
\item Distance for comparison of two frames represented as a point in the 32-dimensional space: Manhattan distance,
\item Threshold for accepting that two frames are a match: $\theta = 0.3$,
\item Matching (pairing) algorithm to determine the number of matches between two summaries: Kannapan algorithm,
\item Measure of similarity between two keyframe summaries: F-measure.
\end{itemize}

Finally, in order to allow for a fair comparison between different summarisation algorithms, we propose the use of $C_U$ as defined in equation (\ref{cu}). Suppose that there are two algorithmic methods giving summaries $P$ and $Q$, respectively. One of them may have a larger F-value for its match to the ground truth (GT) only by virtue of the number of keyframes within. To guard against this, $C_U$ evaluates by how much an algorithm improves over a uniform summary of the same cardinality. Therefore, instead of comparing $F(P,GT)$ with $F(Q,GT)$, we propose to compare
\[
C_U(P) = F(P,GT) - F(U(|P|),GT)
\]
with
\[
C_U(Q) = F(Q,GT) - F(U(|Q|),GT),
\]
where $U(k)$ is a uniform summary with $k$ frames.

If the two rival keyframe summaries $P$ and $Q$ are of the same cardinality, their relative merit can be evaluated by $F(P,GT)$ and $F(Q,GT)$, but the question will remain whether they improve at all on a uniform (or another) baseline.

\begin{figure}
\centering
\includegraphics[width=\linewidth]{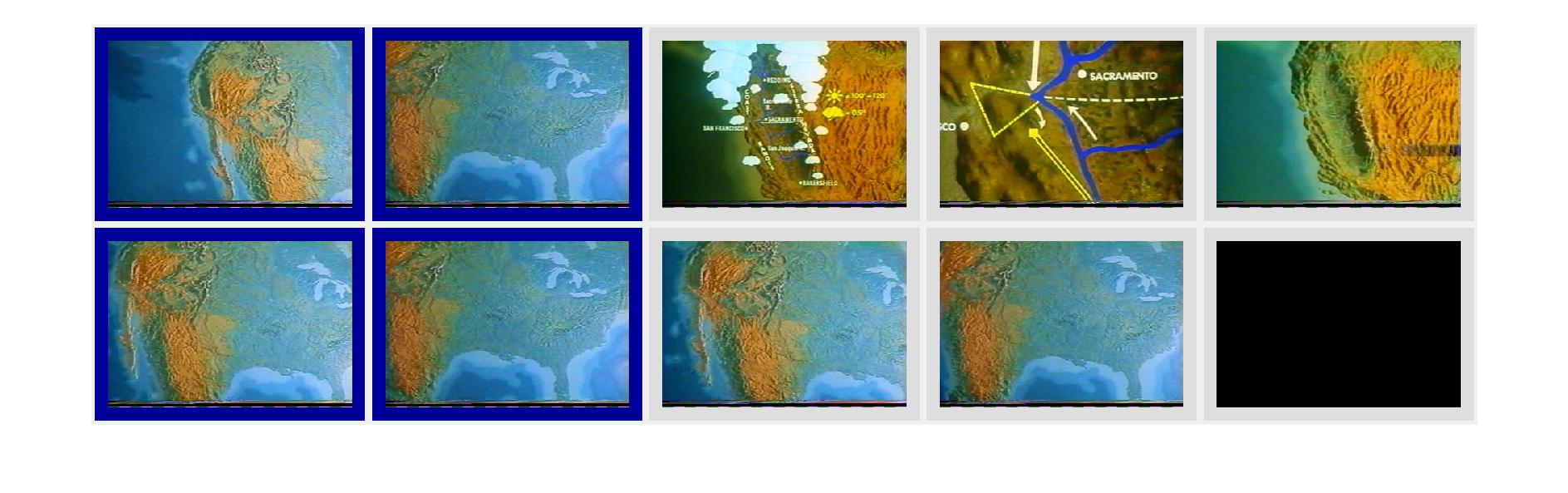}
(a) DT summary: 2 matches
\\ \vspace{5 mm}
\includegraphics[width=\linewidth]{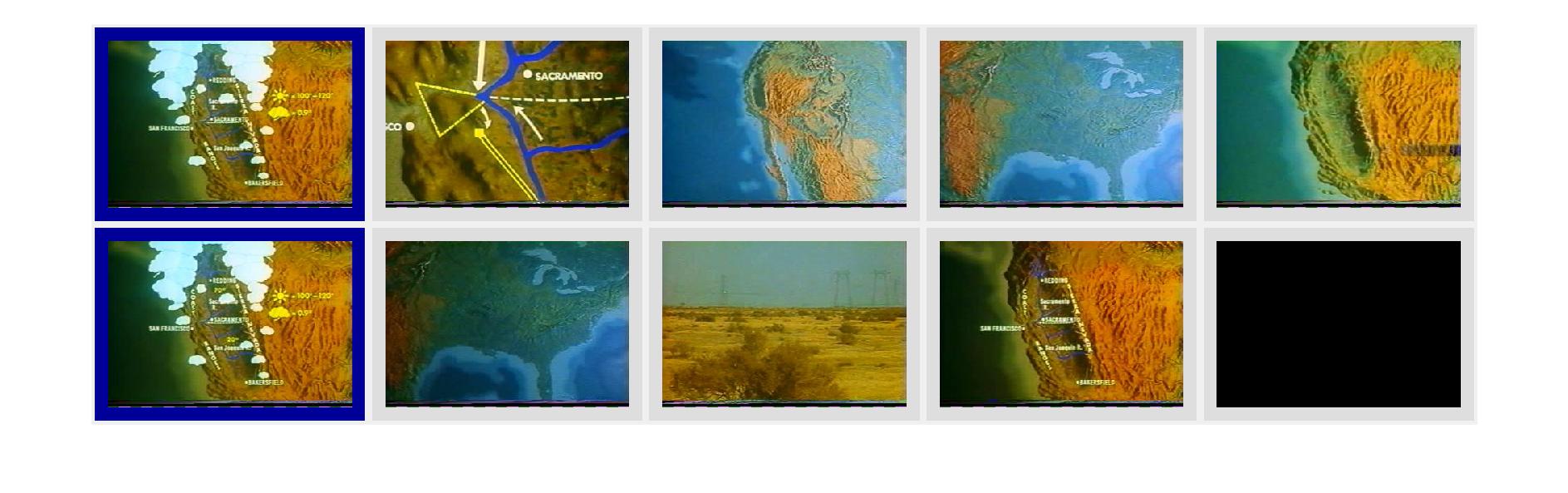}
(b) Uniform summary $U(4)$: one match
\caption{Proposed protocol for video \#22, {\bf DT summarisation method}, user \#3 as a single ground truth.}
\label{fig:proposed1}
\end{figure}

\begin{figure}
\centering
\includegraphics[width=\linewidth]{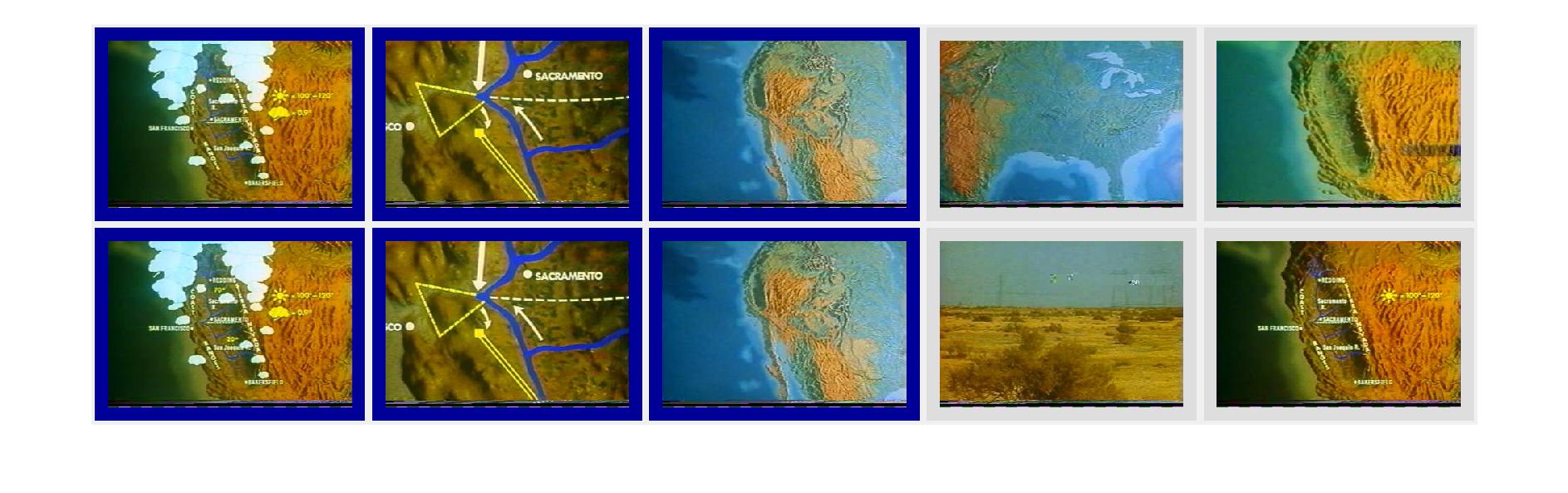}
(a) OV summary: 3 matches
\\ \vspace{5 mm}
\includegraphics[width=\linewidth]{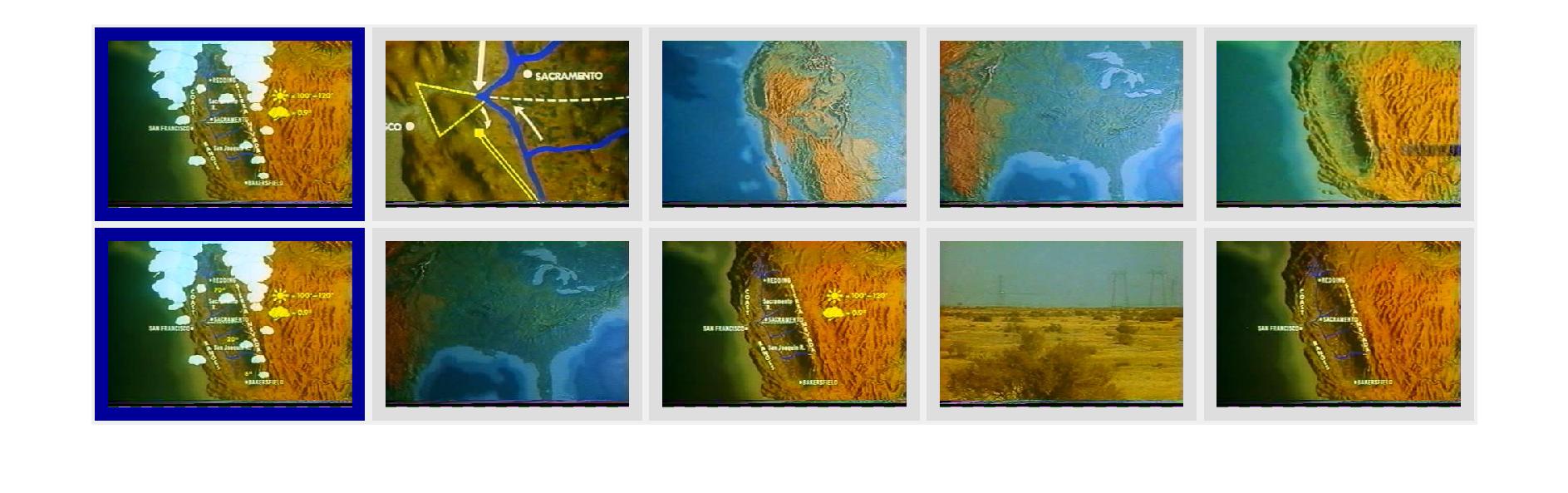}
(b) Uniform summary $U(5)$: one match
\caption{Proposed protocol for video \#22, {\bf OV summarisation method}, user \#3 as a single ground truth.}
\label{fig:proposed2}
\end{figure}

\begin{figure}
\centering
\includegraphics[width=\linewidth]{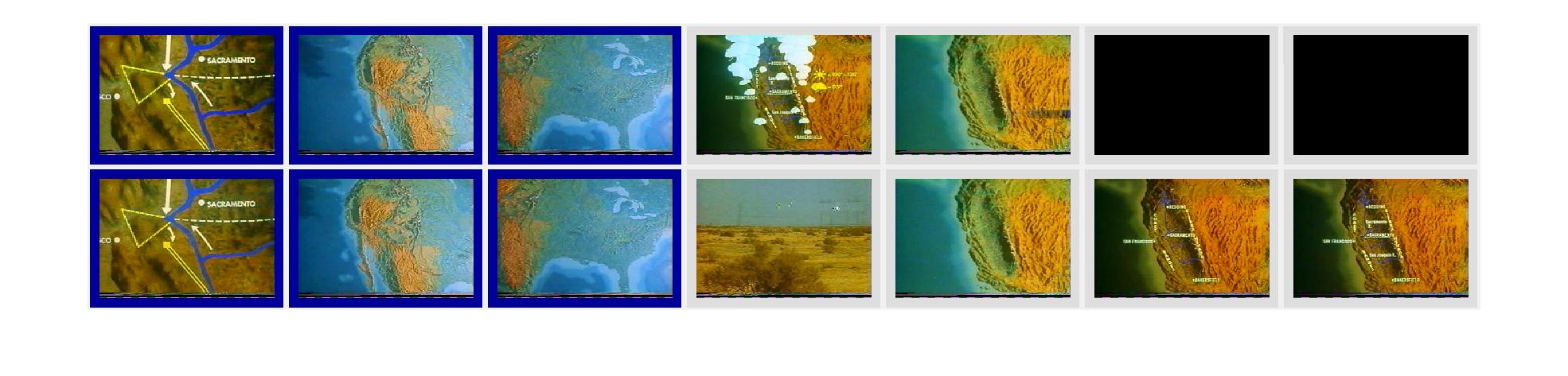}
(a) STIMO summary: 3 matches
\\ \vspace{5 mm}
\includegraphics[width=\linewidth]{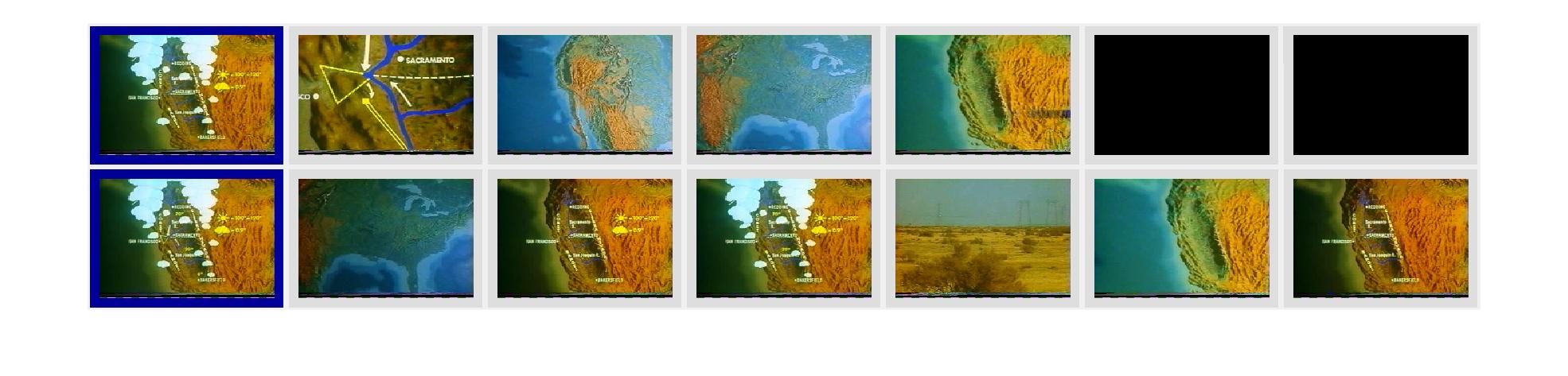}
(b) Uniform summary $U(7)$: one match
\caption{Proposed protocol for video \#22, {\bf STIMO summarisation method}, user \#3 as a single ground truth.}
\label{fig:proposed3}
\end{figure}

\begin{figure}
\centering
\includegraphics[width=\linewidth]{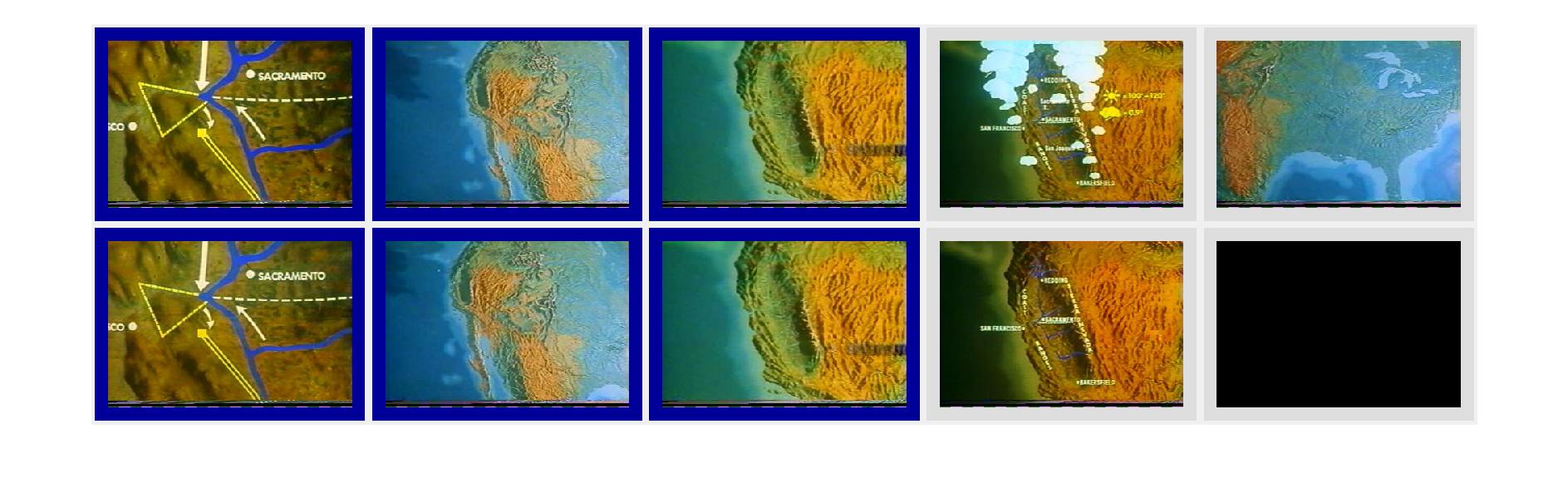}
(a) VSUMM1 summary: 3 matches
\\ \vspace{5 mm}
\includegraphics[width=\linewidth]{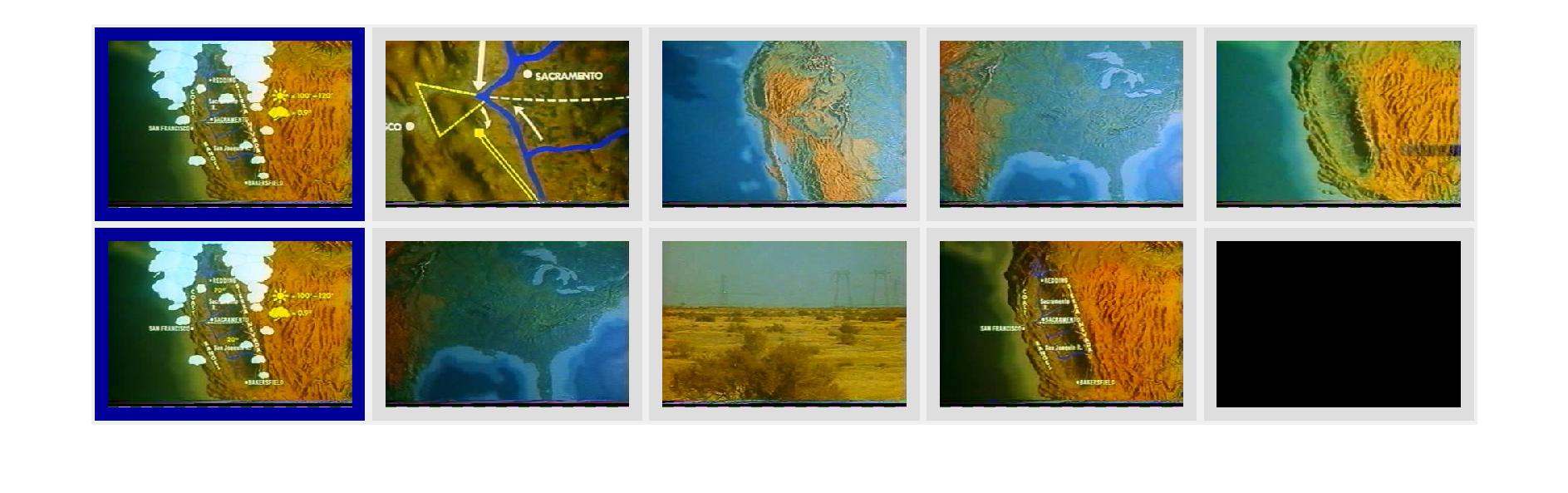}
(b) Uniform summary $U(4)$: one match
\caption{Proposed protocol for video \#22, {\bf VSUMM1 summarisation method}, user \#3 as a single ground truth.}
\label{fig:proposed4}
\end{figure}

\begin{figure}
\centering
\includegraphics[width=\linewidth]{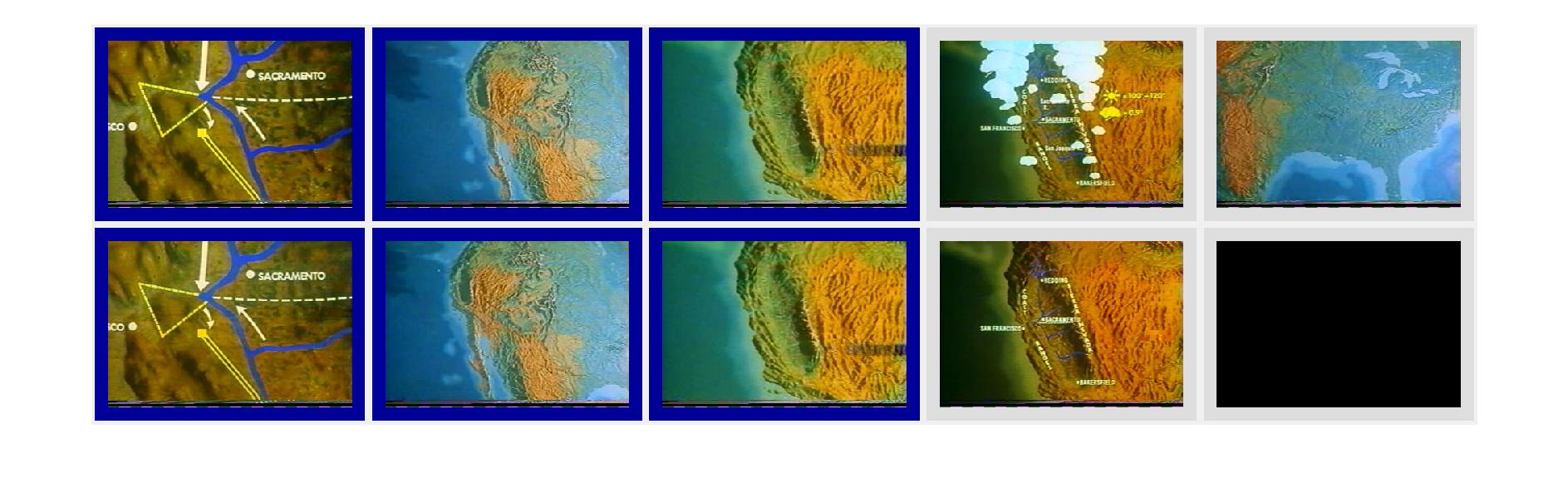}
(a) VSUMM2 summary: 3 matches
\\ \vspace{5 mm}
\includegraphics[width=\linewidth]{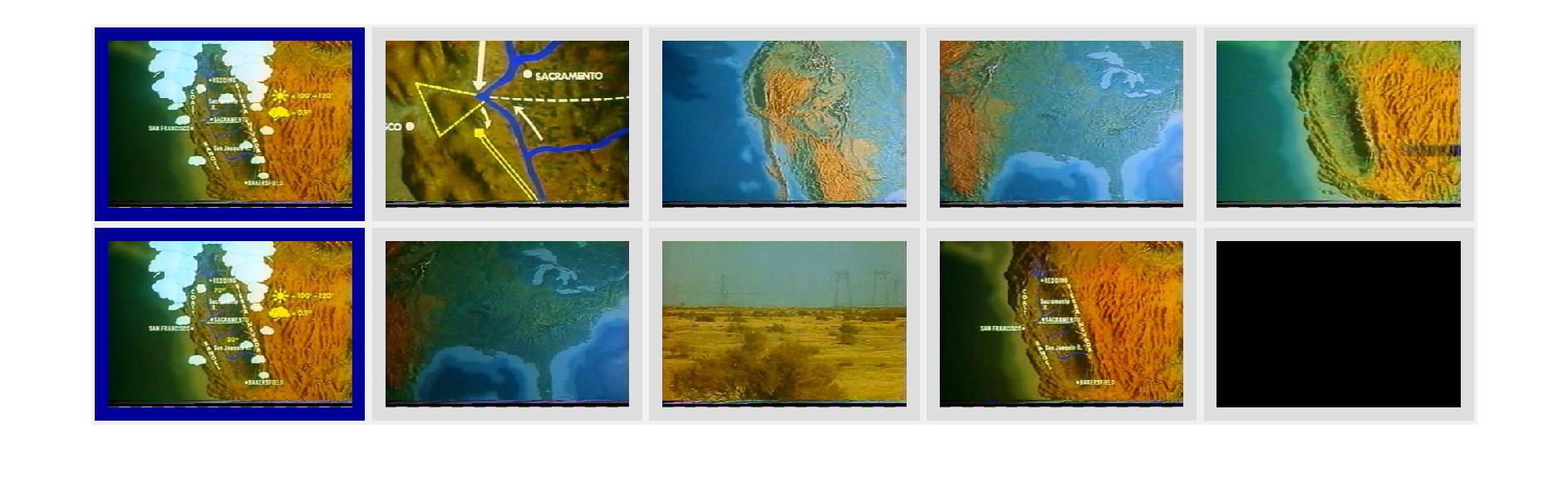}
(b) Uniform summary $U(4)$: one match
\caption{Proposed protocol for video \#22, {\bf VSUMM2 summarisation method}, user \#3 as a single ground truth.}
\label{fig:proposed5}
\end{figure}

We now illustrate how the protocol can be used in practice.\footnote{MATLAB code is provided in GitHub}. Figures~\ref{fig:proposed1} to \ref{fig:proposed5} show the summaries by the 5 algorithmic methods: DT, OV, STIMO, VSUMM1, and VSUMM2, together with the corresponding uniform summary of the same cardinality (the bottom plots). The matches are highlighted with a dark-blue frame. The images in the summaries are arranged so that the matching ones are on the left (recall that we treat the summary as a set, and not as a time sequence). The matches are calculated using the choices of methods and parameters of our proposed protocol. Table~\ref{tab:illu} shows the numerical results for the five methods, assuming that the only available ground truth is the summary of user \#3. (Both the video and the user were chosen at random.)

\begin{table}
\caption{Calculation of the F-values and $C_U$ for the 5 summarisation methods, based on the matches identified by the proposed protocol and illustrated in Figures~\ref{fig:proposed1}--\ref{fig:proposed5}.}
\begin{tabular}{rccc}
Summarisation&&&\\
method $K$&$F(K,GT)$&$F(U(|K|),GT)$&$C_U$\\
\hline
DT&$\frac{2\times 2}{5+4}=0.44$&$\frac{2\times 1}{5+4}=0.22$&$0.22$\\
OV&$\frac{2\times 3}{5+5}=0.60$&$\frac{2\times 1}{5+5}=0.20$&$0.40$\\
STIMO&$\frac{2\times 3}{5+7}=0.50$&$\frac{2\times 1}{5+7}=0.17$&$0.33$\\
VSUMM1&$\frac{2\times 3}{5+4}=0.67$&$\frac{2\times 1}{5+4}=0.22$&$0.45$\\
VSUMM2&$\frac{2\times 3}{5+4}=0.67$&$\frac{2\times 1}{5+4}=0.22$&$0.45$\\
\hline
\end{tabular}
\label{tab:illu}
\end{table}

While in this example the overall ranking of the five summarisation methods is the same according to $F(K,GT)$ and $C_U$, this will not in general be the case. Methods with higher $C_U$ should be preferred. The F-value alone may lead to false claim of matching the ground truth, especially if $F(U(|K|),GT)$ happens to be high. In some cases $C_U$ is negative, which casts a doubt on the validity of the algorithm producing the keyframe summary $K$. 

\section{Conclusion}
\label{sec:con}

We have experimentally investigated a range of choices for different components of a protocol for evaluating the outputs of keyframe-extraction algorithms. A new measure called ``discrimination capacity'' $C_U$ is proposed, which evaluates by how much a given summary improves on the uniform keyframe summary of the same cardinality. Using $C_U$ and the VSUMM video collection, we offer empirical recommendations, and propose a full protocol for comparison of keyframe summaries, listed at the start of sub-section \ref{proposed}. 

We discovered that the most acclaimed feature spaces such as CNN and SURF are not the best choices for our protocol. A 32-bin hue histogram feature space fared better than the high-level features. Our study also contains a comprehensive collection of algorithms for matching (pairing) two summaries of different cardinalities. These algorithms did not make a profound difference to the output, therefore we chose a simple, yet efficient matching algorithm published recently~\cite{Kannappan16}. 

Our future work will include looking into semantic comparisons between frames and summaries in addition to matching based solely on visual appearance. Combinations thereof as well as incorporating the time tag in the comparisons will be explored.

\section*{Acknowledgment}
This work was done under project RPG-2015-188 funded by The Leverhulme Trust, UK.

\section*{References}


\end{document}